\documentclass[runningheads]{llncs}
\usepackage{graphicx}

\usepackage{tikz}
\usepackage{comment}
\usepackage{amsmath,amssymb} 
\usepackage{color}

\usepackage[accsupp]{axessibility}  

\usepackage{booktabs}
\usepackage{multirow}
\usepackage{listings}
\usepackage{xcolor}
\usepackage{bbm}
\usepackage{subfig}
\usepackage{graphicx}
\usepackage{graphbox}

\usepackage{array}
\newcolumntype{P}[1]{>{\centering\arraybackslash}p{#1}}

\DeclareRobustCommand\onedot{\futurelet\@let@token\@onedot}
\def\@onedot{\ifx\@let@token.\else.\null\fi\xspace}
\def\eg{\emph{e.g}., } 
\def\ie{\emph{i.e}., }

\usepackage[pagebackref,breaklinks,colorlinks]{hyperref}

\usepackage[capitalize]{cleveref}
\crefname{section}{Sec.}{Secs.}
\crefname{section}{Section}{Sections}
\crefname{table}{Table}{Tables}
\crefname{table}{Tab.}{Tabs.}

\begin{document}
\pagestyle{headings}
\mainmatter
\def\ECCVSubNumber{5010}  

\title{Semi-supervised Object Detection via Virtual Category Learning} 


%
\author{Changrui Chen\inst{1} \and
Kurt Debattista\inst{1} \and
Jungong Han\inst{1,2}\thanks{corresponding author} }
\institute{University of Warwick, WMG \\
\email{\{Changrui.Chen; K.Debattista\}@warwick.ac.uk}
\and
Aberystwyth University, Computer Science \\
\email{juh22@aber.ac.uk}
}

\maketitle

\begin{abstract}

Due to the costliness of labelled data in real-world applications, semi-supervised object detectors, underpinned by pseudo labelling, are appealing. However, handling confusing samples is nontrivial: discarding valuable confusing samples would compromise the model generalisation while using them for training would exacerbate the confirmation bias issue caused by inevitable mislabelling. To solve this problem, this paper proposes to use confusing samples proactively without label correction. Specifically, a virtual category (VC) is assigned to each confusing sample such that they can safely contribute to the model optimisation even without a concrete label. It is attributed to specifying the embedding distance between the training sample and the virtual category as the lower bound of the inter-class distance. Moreover, we also modify the localisation loss to allow high-quality boundaries for location regression. Extensive experiments demonstrate that the proposed VC learning significantly surpasses the state-of-the-art, especially with small amounts of available labels.

\vspace{-0.5em}
\end{abstract}

\section{Introduction}
\label{sec:intro}

The Deep Learning community is suffering from the expensive labelling cost of large-scale datasets. Semi-supervised learning, which makes use of limited labelled data combined with large amounts of unlabelled data for training, has shown great potential to reduce the reliance on large amounts of labels~\cite{Yen-Cheng_2021_Unb,David_2019_Mix,Kihyuk_2020_Fix,Junnan_2020_CoM}. In particular, this work focuses on object detection, a task that typically requires a significant effort to label, under the semi-supervised setting, especially with extremely small amounts of labelled data.

\begin{figure}[t]
\centering
\includegraphics[align=c,width=0.36\columnwidth]{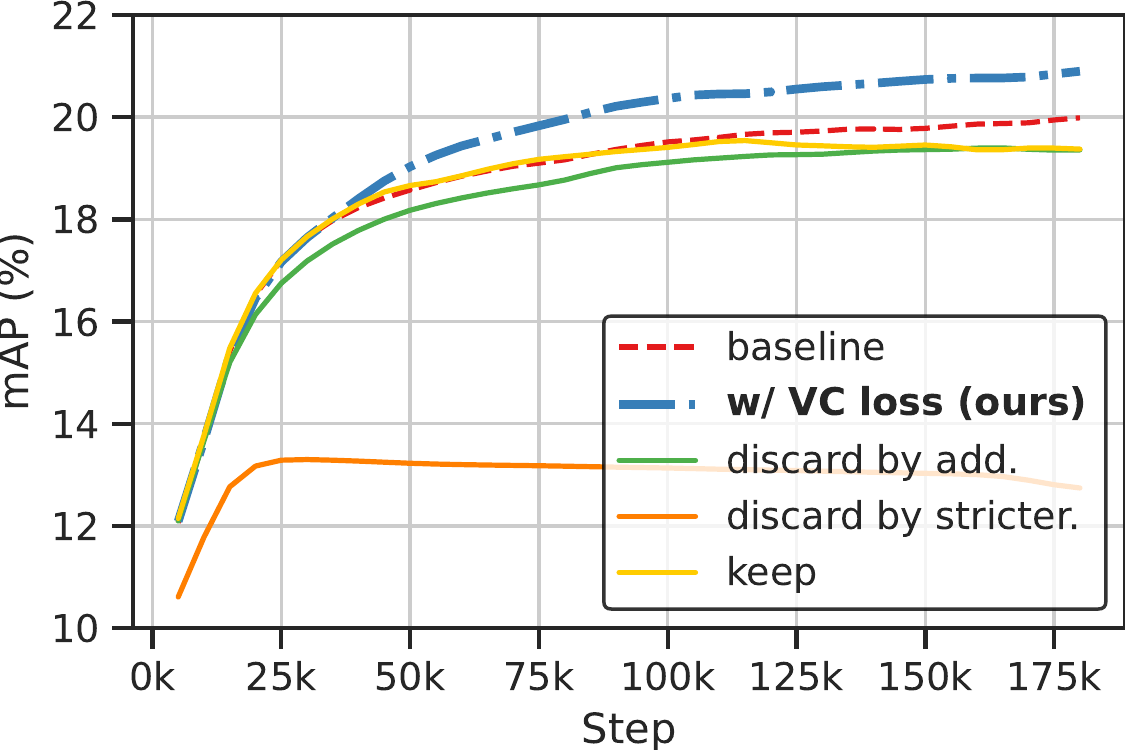}
\hspace{2em}
\includegraphics[align=c,width=0.45\columnwidth]{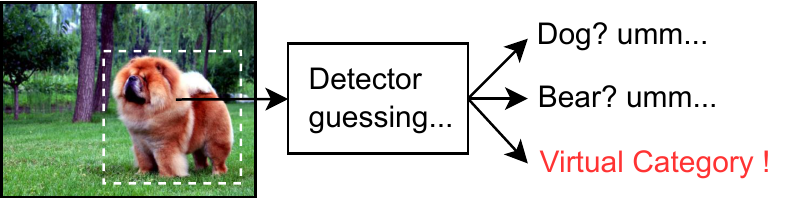}
\vspace{-5px}
\caption{Left: The mAP of a semi-supervised detector~\cite{Yen-Cheng_2021_Unb} with a preset confidence score filtering on 1\% labelled MS COCO~\cite{Tsung-Yi_2014_Mic}. The mAP sees a decrease with all strategies for dealing with confusing samples (\eg the bear-like dog at the right) except for our VC learning. The additional filtering mechanism (add.) is the temporal stability verification proposed in this paper. Stricter filtering (stricter.) is simply raising the threshold in the score filtering from 0.7 to 0.8. Right: Illustration of the basic idea of our virtual category.}\label{fig:intro}
\vspace{-2em}
\end{figure}

Pseudo labelling~\cite{Lee_2013_Pse}, one of the most advanced methods in semi-supervised learning, has been recently introduced to semi-supervised object detection (SSOD) \cite{Yen-Cheng_2021_Unb,Qiang_2021_Ins,Qize_2021_Int,Mengde_2021_End}. Unlabelled data are automatically annotated by the detector itself~\cite{Kihyuk_2020_SATC} (or an exponential moving average version~\cite{Yen-Cheng_2021_Unb}) and then fed back to re-optimise the detector. Typically, due to the limited diversity of samples in a extremely small available set of labelled data, the detector is usually unable to make a confident decision on some unseen confusing samples when inferring their pseudo labels. Two strategies are usually adopted in SSOD to deal with the confusing samples: a) discarding all of them by a strict filtering mechanism~\cite{Kihyuk_2020_SATC}, or b) retaining them with all potential labels~\cite{Qize_2021_Int}. However, none of these two options is optimal. Admittedly, the value of confusing hard samples is clear to see, since hard example mining~\cite{Abhinav_2016_Tra} has successfully proven its effectiveness in fully-supervised learning. If all confusing samples are rejected by a strict filtering mechanism in SSOD, their positive contributions will be wasted, while the remaining well-fitted samples only marginally contribute to performance improvements. On the contrary, simply keeping them all does not work either due to the involvement of too many incorrect pseudo labels. Arbitrarily optimising semi-supervised detectors with these noisy labels will lead to the confirmation bias issue~\cite{Eric_2020_Pse}. To demonstrate these points, we show in~\cref{fig:intro} the mean Average Precision (mAP) of a semi-supervised detector with different strategies on 1\% labelled MS COCO~\cite{Tsung-Yi_2014_Mic}. We observe noticeable performance degradations when either choosing one stricter filtering mechanism (orange line) or adding an additional filtering (green line) to reject confusing samples. Likewise, simply keeping all pseudo labels of confusing samples (yellow line) also ends up with a decreased mAP since the unreliable pseudo labels aggravate the confirmation bias issue and even give rise to training collapse.

Therefore, efforts are dedicated to exploring how to correct the biased pseudo labels to utilise the confusing samples without any risks. Existing methods~\cite{Xinshao_2021_Pro,Junnan_2020_CoM} initially investigate relatively easy tasks such as classification on CIFAR~\cite{krizhevsky2009learning}. 
However, progress has not yet been made for a complex task, such as object detection with extremely small amounts of labelled data, due to its difficulty in correcting object pseudo labels. Then, a question arises: \textit{what if we do not discard any confusing samples but consider their contributions, which may not necessarily need the concrete label information, during the model training?} This paper answers this question by proposing a novel Virtual Category (VC) learning, based on an observation that the confusing samples with doubtful pseudo labels can be profitable for training. 

Specifically, we discover that building a potential category (PC) set consisting of the possible categories of a confusing sample $x$, compared to determining the exactly correct label for it, seems more feasible. Therefore, instead of selecting the correct one from the PC set, which is usually hard, we compromise by proposing a virtual category label to \textit{take the place of all unreliable labels in the PC set}. As shown in the example of the right part of \cref{fig:intro}, the PC set of this object consists of two categories (\ie dog and bear). A new modified Cross Entropy loss function, namely VC loss, allows the detector to be optimised with the virtual category label. By doing so, ignoring the categories in the PC set will disable the corresponding output logits, thus avoiding any wrong gradients to mislead the model optimisation. Most importantly, the proposed virtual category specifies a reasonable lower bound for the inter-class distance. Hence, the decision boundary can consistently benefit from the confusing data without suffering from the confirmation bias issue. With regards to the PC set, we investigate multiple methods to build it, including temporal stability verification and cross-model verification.  As can be seen in~\cref{fig:intro}, the mAP of the model armed with the proposed VC loss (dot-dashed line) sees a significant increase due to the effective use of the confusing samples. Furthermore, we also enhance localisation training by decoupling the horizontal and vertical location qualities of a pseudo bounding box, which enables us to use the high-quality boundaries to optimise the bounding box regression head.  

We apply our VC learning to a semi-supervised detector and evaluate it on MS COCO~\cite{Tsung-Yi_2014_Mic} and Pascal VOC~\cite{Mark_2015_The} datasets with extremely low label ratios, including 0.5\%, 1\%, 2\%, 5\% and 10\%. On MS COCO, our method achieves 19.46 mAP with only 586 labelled images; this outperforms some recently published semi-supervised detectors~\cite{Kihyuk_2020_SATC,Qiang_2021_Ins} with 1000+ labelled images. The contributions of this paper are summarised as follows:

\begin{itemize}
\item We propose VC learning to alleviate the confirmation bias issue caused by confusing samples in SSOD. The proposed method takes the initiative to use confusing samples with unreliable pseudo labels for the first time. 
\item A detailed explanation of the VC learning is provided to demonstrate the mathematical feasibility and intuitive effect. Our intriguing  findings highlight the need to rethink the usage of confusing samples in SSOD tasks.
\item We demonstrate how the proposed method surpasses state-of-the-arts by a significant margin on two benchmark datasets with different label ratios.
\end{itemize}

\section{Related Works}

\textbf{Object Detection,} which finds significant applications in downstream tasks, aims to distinguish foreground objects in images or videos and identify them. Object detectors so far can be generally divided into three types: 1) Two-stage detectors~\cite{Shaoqing_2017_Fas,Zhaowei_2018_Cas},  represented by Faster RCNN~\cite{Shaoqing_2017_Fas}; 2) One-stage detectors~\cite{Joseph_2016_You,Alexey_2020_YOL,Tsung-Yi_2020_Foc,Nicolas_2020_End}, such as the YOLO series~\cite{Joseph_2016_You,Alexey_2020_YOL}; and 3) Point-based Detectors~\cite{Kaiwen_2019_Cen,Zhi_2019_FCO,law2018cornernet}, such as Center Net~\cite{Kaiwen_2019_Cen}. The main difference between two-stage and one-stage detectors lies in whether an additional module is used to generate candidate region proposals for classification and localisation. Point-based detectors discard anchor boxes and instead use points and sizes to represent objects. In this paper, Faster RCNN, one of the most widely used detectors, serves as our baseline model.

\vspace{0.5em}\noindent\textbf{Semi-supervised Learning} is a training scheme that uses only a very small amount of labelled data and a large amount of unlabelled data to train a machine learning model. Recently proposed semi-supervised methods mainly focus on image classification~\cite{Kihyuk_2020_Fix,David_2019_Mix}. Consistency regularisation is one of the most advanced methods. It requires models to produce consistent outputs when the inputs are perturbed. Image augmentations, such as flipping, Cutout~\cite{Terrance_2017_Imp}, or Gaussian Blurring, are usually applied to perturb inputs~\cite{Samuli_2017_Tem,Philip_2014_Lea}. Some solutions take advantage of adversarial learning and proposed learnable adversarial augmentations~\cite{Takeru_2019_Vir}. Inspired by the noisy label learning method~\cite{kim2019nlnl}, negative sampling~\cite{chen2020negative} improved the performance of semi-supervised algorithms by randomly sampling a non-target label for an unlabelled data. Alternatively, Pseudo labelling~\cite{Lee_2013_Pse}, which is used in this paper, is another semi-supervised technique that has proven successful. It minimises the entropy of the predictions of unlabelled data by generating pseudo labels for them.

\vspace{0.5em}\noindent\textbf{Semi-supervised Object Detection (SSOD)} originates from semi-supervised classification, where only a small amount of bounding box labelled data and numerous unlabelled data are available for training a detector. Most of the recently proposed SSOD methods can be broadly categorised into two different types: 1) consistency-based and 2) self-supervised (pseudo-labelling) methods. Consistency-based semi-supervised object detectors, such as CSD-SSD~\cite{Jisoo_2019_Con}, when dealing with unlabelled images, apply consistency regularisation on the predicted classification probability vectors and regression vectors of the input image and its mirror version. Alternatively, self-supervised detectors~\cite{Yen-Cheng_2021_Unb,Qiang_2021_Ins,Qize_2021_Int,Kihyuk_2020_SATC} were inspired by the pseudo-labelling classifier, which provides detectors with task-related supervision information. Although the self-supervised detectors show good potential, they are struggling with the confirmation bias issue when training with confusing samples. This paper alleviates this issue via a novel VC learning.

\section{Methodology}
In this section, the overall problem is first defined. The VC learning and its explanation are subsequently described.

\subsection{Problem Definition} 
\begin{figure*}[t]
  \centering
   \includegraphics[width=0.95\textwidth]{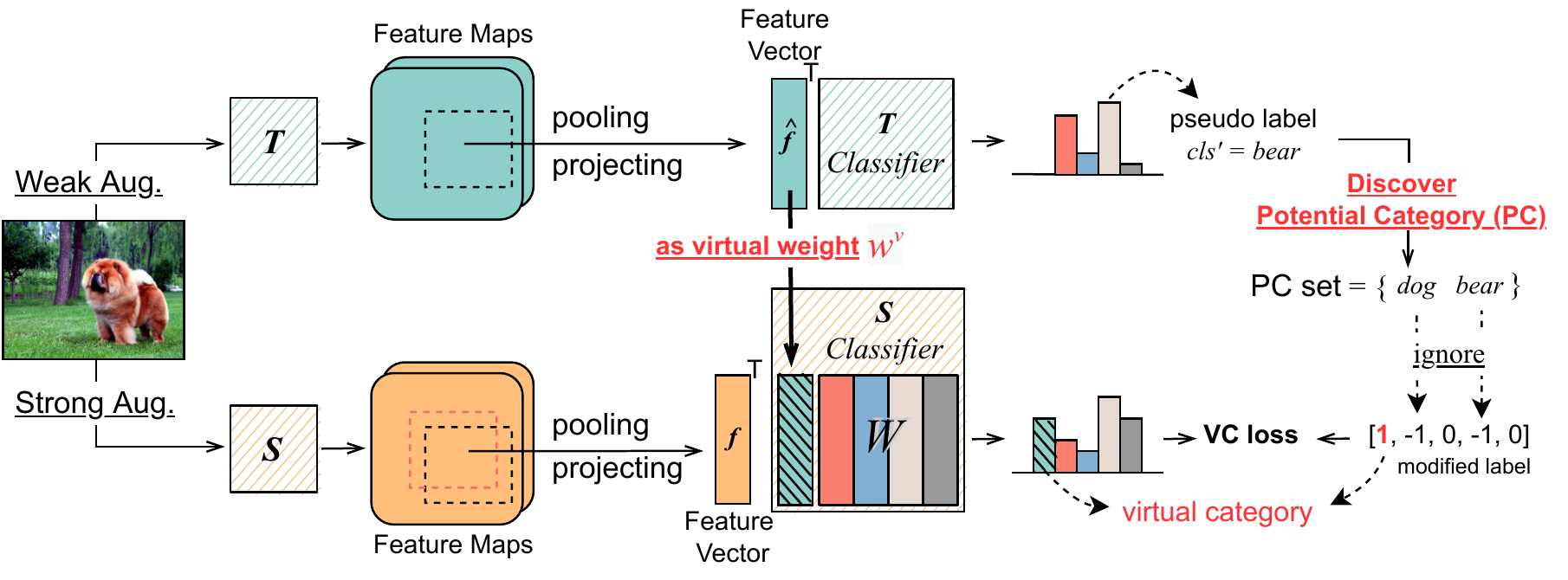}
   \vspace{-1.2em}
   \caption{The pipeline of the proposed VC learning when dealing with a confusing sample. The regression branch is ignored for simplicity. $T$ is the teacher model. $S$ represents the student model. The box with the black dashed line on feature maps indicates the position of the pseudo label. When training the student classifier with a RPN-generated proposal box (pink dashed box in the figure) assigned to this pseudo label, the weight matrix $W$ of the student classifier is extended by a virtual weight $w^v$, which is transformed from the teacher feature vector $\hat f$ of the pseudo label area.}
   \label{fig:framework}
\vspace{-1em}
\end{figure*}

In the problem of SSOD, two data subsets $\mathcal{D}^l$ and $\mathcal{D}^u$ are  given for model optimisation, where $\mathcal{D}^l = \{(x^l_n, b^l_n) |_{n=0}^{N^l}\}$ is the subset with available ground truth bounding box $b^l$, $\mathcal{D}^u = \{x^u_n |_{n=0}^{N^u}\}$ is the unlabelled subset. $N^l$ and $N^u$ are the number of labelled and unlabelled data. Usually, $N^u \gg N^l$. $b=[x_1, y_1, x_2, y_2, cls]$ is the label -- bounding box, where the first 4 numbers indicate the coordinates of the top-left and bottom-right points and $cls$ is the index of the category label. This paper follows the self-supervised framework with a score filtering mechanism \cite{Yen-Cheng_2021_Unb} to generate pseudo labels $b'$ of $x^u$ for re-optimising. As shown in~\cref{fig:framework}, two detectors $T$ and $S$ sharing the same architecture are introduced. The parameters of the teacher detector $T$ are updated by the parameters of the student $S$ with a momentum.

\subsection{Virtual Category Learning}
\label{sec:VCloss}

In \cref{fig:framework}, the teacher classifier produces the categorical probability of the candidate object (black dashed box in \cref{fig:framework}) according to the pooled and projected feature vector $\hat f$ of it. Typically, the category with the highest probability, \eg $cls'=bear$ here, will then be used as the pseudo label of the candidate proposal box assigned to this instance during the student training. However, incorrect pseudo labels may mislead the training.

In this paper, we propose VC learning which modifies the pseudo category label with an additional virtual category to allow the student model to be optimised safely by confusing samples. Once the initial pseudo label $cls'=bear$ is obtained, a potential category (PC) discovery operation is performed to construct a PC set $\{dog, bear\}$ for this training sample. We find that PC discovery is relatively feasible compared to designing a correction function: $g(cls' | f)=GT$ , especially when the labelled subset $\mathcal{D}^l$ is much smaller than $\mathcal{D}^u$. The discovery function will be introduced in the following \cref{sec:pcset}.

If the PC set contains more than one category, it means that this sample is confusing to the detector. In \cref{fig:framework}, the bear-like dog is a confusing sample with $PCset=\{dog, bear\}$. To allow all the proposal boxes (\eg pink dashed box in \cref{fig:framework}) assigned to this bear-like dog consistently contribute to the optimisation of the student detector $S$ rather than arbitrarily discarding them, the weight matrix $W$ in the student classifier is extended by an additional weight vector $w^v$ named \textit{virtual weight}. The virtual weight $w^v$ is obtained from the feature vector $\hat f$, which is the feature of this confusing sample in the teacher model. Thus, the virtual weights for different confusing samples are different. The normalisation and scaling are applied to $\hat f$ to ensure the norm of the  produced $w^v$ and the norm of the weight vectors in $W$ are in the same range. We investigate two scaling factors, \ie a constant value and an adaptive factor calculated by averaging the norm of all the weight vectors in $W$. With the extended weight matrix, the size of the student classifier output (\ie logits) is therefore increased by 1:
\begin{equation}
	f^\top \cdot [\overbrace{w^v,w^{c_0},...,w^{c_{N-1}}}^{1+N}]=[\overbrace{l^v,l^{c_0},...,l^{c_{N-1}}}^{1+N}],
\label{eq:vcfw}
\end{equation} where $N$ is the number of the predefined categories, $l^v$ and $l^{c_i}$ are the logits of the virtual category and the class $c_i$ respectively. To calculate the loss value of the extended logits, the one-hot label is modified by providing a positive label `1' for the virtual category. Thanks to the virtual category taking on the responsibility of being the target category, the confusing categories in the PC set can be ignored (indicated by `-1' in \cref{fig:framework}), thereby avoiding any possible misleading as we can hardly determine which one is the real ground truth.

\begin{figure}[t]
\centering
\vspace{-1.5em}
\subfloat[]{
  \centering
   \includegraphics[width=0.45\linewidth,height=0.18\linewidth]{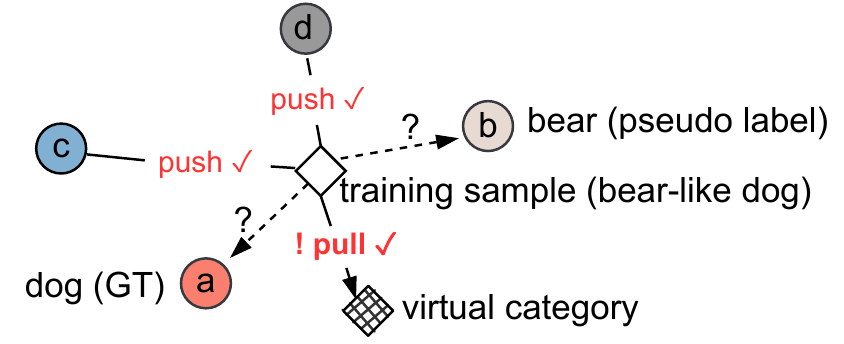}
   \label{fig:explain}
}
\subfloat[]{
    \centering
    \includegraphics[width=0.37\textwidth,height=0.18\linewidth]{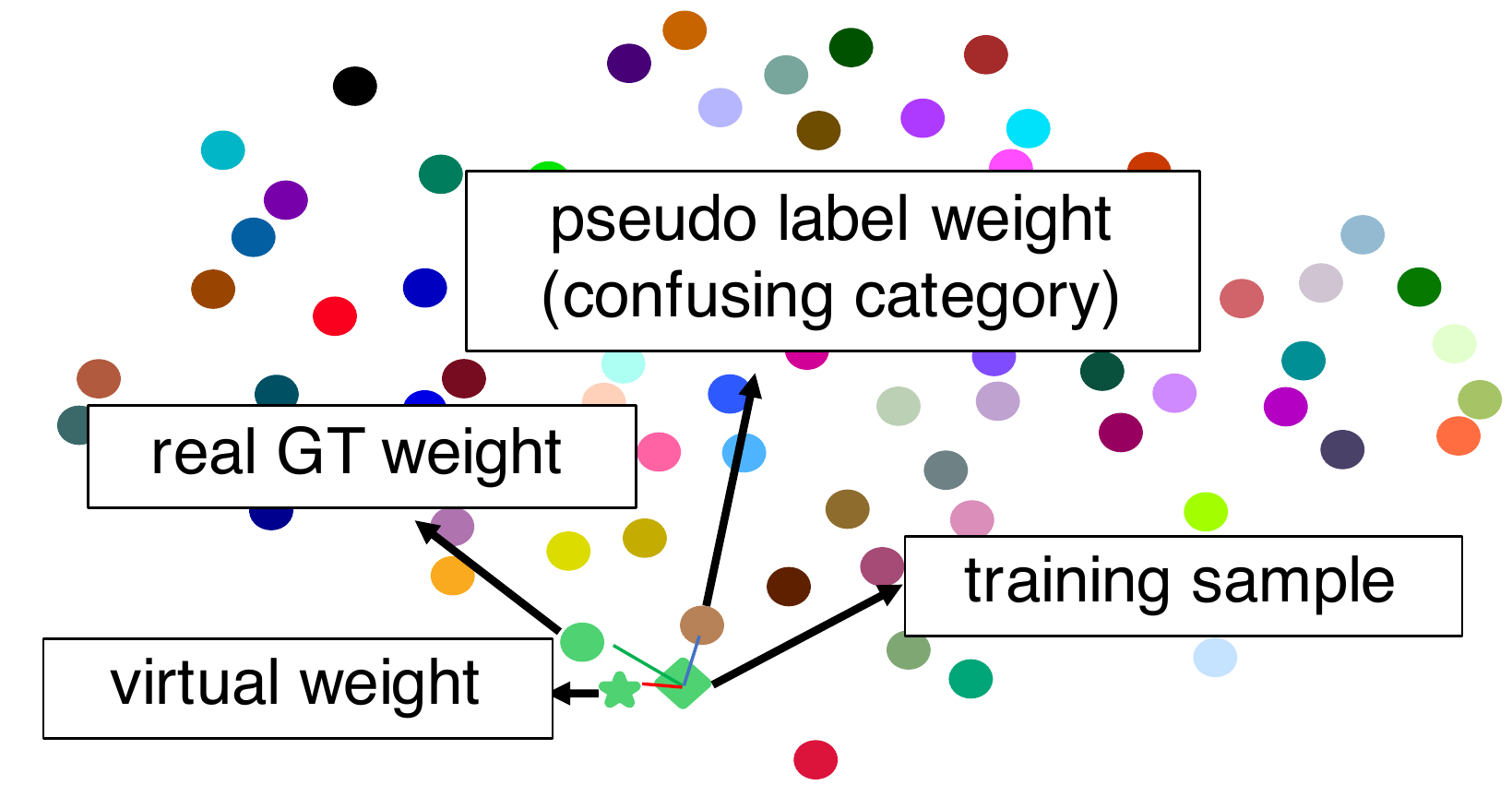}
    \label{fig:ablation-tsne}
}
\vspace{-1em}
\caption{a) Explanation of VC loss in the feature space. The circles are the embeddings of the category cluster centres. The hollow diamond is the embedding of the training sample. The cross-hatch diamond is the embedding of the virtual category. b) T-SNE visualisation of the virtual weight (star), learned classifier weights (circles), and a training sample (diamond).}
\vspace{-1em}
\end{figure}

\vspace{0.5em}\noindent\textbf{Explanation} How the proposed VC learning works can be interpreted from both the feature space aspect and the mathematical feasibility.

In the feature space, as shown in~\cref{fig:explain}, let the circles indicate the centres of the predefined categories. Pulling the training sample (diamond) to the circles $a$ or $b$ is risky since we don't know which one is the real ground truth. Given the virtual category, the decision boundary can still be optimised with VC learning, as the virtual category providing a safe optimising direction: pushing the training sample away from the circles $c$ and $d$ and pulling it closer to the diamond of the virtual category. Although one may suspect that our approach looks similar to the contrastive learning~\cite{Kaiming_2020_Mom,Ting_2020_A_S} in terms of the optimisation target, they differ in several aspects. Firstly, contrastive learning operates before the task-relevant layer (\ie the classifier). As a result, it only drives the backbone to extract better features but contributes nothing to the task-relevant layer. While our approach acts after the classifier so that the gradient of virtual category can backpropagate to not only the backbone but also the weight vectors in the classifier. Secondly, the weight vectors of the other categories in the classifier naturally constitute negative samples such that there is no need to maintain a negative sample pool, which has been a worrying bottleneck for contrastive learning. \cref{fig:ablation-tsne} presents the T-SNE~\cite{Van_2008_Vis} visualisation of the weight vectors $w^{c_i}$ in the trained classifier, and the virtual weight $w^v$ and the feature $f$ of a sample. Our virtual category is clearly a better training target compared to the pseudo labels, which is in line with the feature space aspect explanation.

To explain our method from the mathematical aspect, we first define the VC loss starting from the cross entropy~(CE) loss. We abbreviate the category $c_i$ by its index $i$ in the following. Assuming a batch size of 1, the CE loss is:
\begin{equation}
    \label{eq:celoss}
    \mathcal{L}_{CE} = -log(\frac{e^{f^\top \cdot w^{i=GT}}}{\sum_{i=0}^{N} e^{f^\top \cdot w^{i}}}) = log(\sum_{i=0}^{N}e^{l^{i}-l^{GT}}),
\end{equation}
\noindent where $f \in \mathbb{R}^{channel \times 1}$ is the input feature vector of the last linear layer (\ie classifier), $w^{i} \in \mathbb{R}^{channel\times 1}$ is the corresponding weight vector of the category $c_{i}$ in the last linear layer, $l^i=f^\top \cdot  w^i$ is the logit of the category $c_{i}$, $N$ is the number of the predefined categories, and $GT$ is the index of the ground truth. Here, we ignore the bias in the last linear layer for simplicity. With \cref{eq:vcfw}, the VC loss can be defined as follows according to \cref{eq:celoss}:
\begin{equation}
    \label{eq:vcloss}
    \mathcal{L}_{VC} = log(\sum_{i=0,i \notin PC}^{N+1}e^{l^{i}-l^{v}}),
\end{equation} where $PC$ is the PC set. When $i=N$, the index indicates the virtual category. For example, $l^{i=N}=l^{v}$. $i \notin PC$ means the uncertain potential categories in the PC set are ignored. 

The intuitive target of minimising CE loss is to get a large logit $l^{GT}$ of the ground truth and small values for the rest of the categories $l^{i \neq GT}$. In a self-supervised scheme, the pseudo classification label $cls'$ of a unlabelled data $x_u$ is generated by the model itself or the EMA (Exponential Moving Average) model. As~\cref{eq:celoss} can be a smooth approximation of the \textit{max} function~\cite{boyd2004convex} (explanation without the approximation and more details are in the supplementary document), the CE loss with the pseudo label $cls'$ can be expressed as:
\begin{equation}
    \label{eq:celoss-2}
    \mathcal{L}_{CE} = log(\sum_{i=0}^{N}e^{l^{i}-l^{cls'}}) \approx max_{i \in \{0, ..., N-1\}}(l^i-l^{cls'}),
\end{equation} minimising \cref{eq:celoss-2} is expected to satisfy:
\begin{equation}
    \label{eq:celoss-expect}
(l^{i \neq cls'} - l^{cls'}) \leq (l^{i=cls'} - l^{cls'}) \triangleq 0, \hspace{2em} i.e., \hspace{2em} l^{i \neq cls'} \leq l^{cls'}.
\end{equation}

\noindent Since $cls'$ may not always be correct, when $cls'$ is not equal to the real ground truth $GT$, satisfying~\cref{eq:celoss-expect} leads to $l^{GT} \leq l^{cls'}$, thereby aggravating the issue typically termed confirmation bias.

However, for VC loss, following the derivation of~\cref{eq:celoss-2}, we get:
\begin{equation}
    \label{eq:virtualloss-1}
    \mathcal{L}_{VC} = log(\sum_{i=0,i \notin PC}^{N+1}e^{l^{i}-l^{v}}) \approx max_{i \in \{0, ..., N\} \backslash PC}(l^i-l^{v}).
\end{equation} Similar to~\cref{eq:celoss-expect}, minimising~\cref{eq:virtualloss-1} is expecting:
\begin{equation}
    \label{eq:vcloss-expect}
(l^{i \neq v \wedge i \notin PC } - l^{v}) \leq (l^{i=v} - l^{v}) \triangleq 0, \hspace{2em} i.e., \hspace{2em} l^{i \neq v \wedge i \notin PC} \leq l^{v}.
\end{equation} Comparing~\cref{eq:vcloss-expect} with~\cref{eq:celoss-expect} reveals:

\begin{enumerate}
	\item $\mathcal{L}_{VC}$ firstly ignores the logits $l^{i \in PC}$ of the confusing categories in the PC set when satisfying the inequation in \cref{eq:vcloss-expect}, thereby avoiding misleading the training.
	\item Additionally, it provides an alternative upper bound $l^{v}=f^\top \cdot w^{v}$ for all the rest logits $l^{i \notin PC}$. As before-mentioned, $w^{v}$ is the normalised and scaled feature vector $\hat f$ according to the length (norm of the vector) of the weight vectors $w^i$. The directions of $w^{v}$ and all the $w^i$ can represent the information of different categories~\cite{Hang_2018_Low}. The $l^{v}=f^\top \cdot w^{v}$ should be larger since $w^v$ are obtained by $\hat f$ which is the feature in the teacher of the exactly same object of feature $f$. The shared information between $f$ and $\hat f$ should be the most. Thus, $l^{v}$ can be a meaningful upper bound for all the rest logits $l^{i \notin PC}$.
\end{enumerate}

\subsection{Potential Category Set}
\label{sec:pcset}
Two methods are explored in this paper to discover the potential category.

\vspace{0.5em}\noindent\textit{a. Temporal stability}

\begin{figure}[t]
  \centering
   \includegraphics[align=c,width=0.3\textwidth]{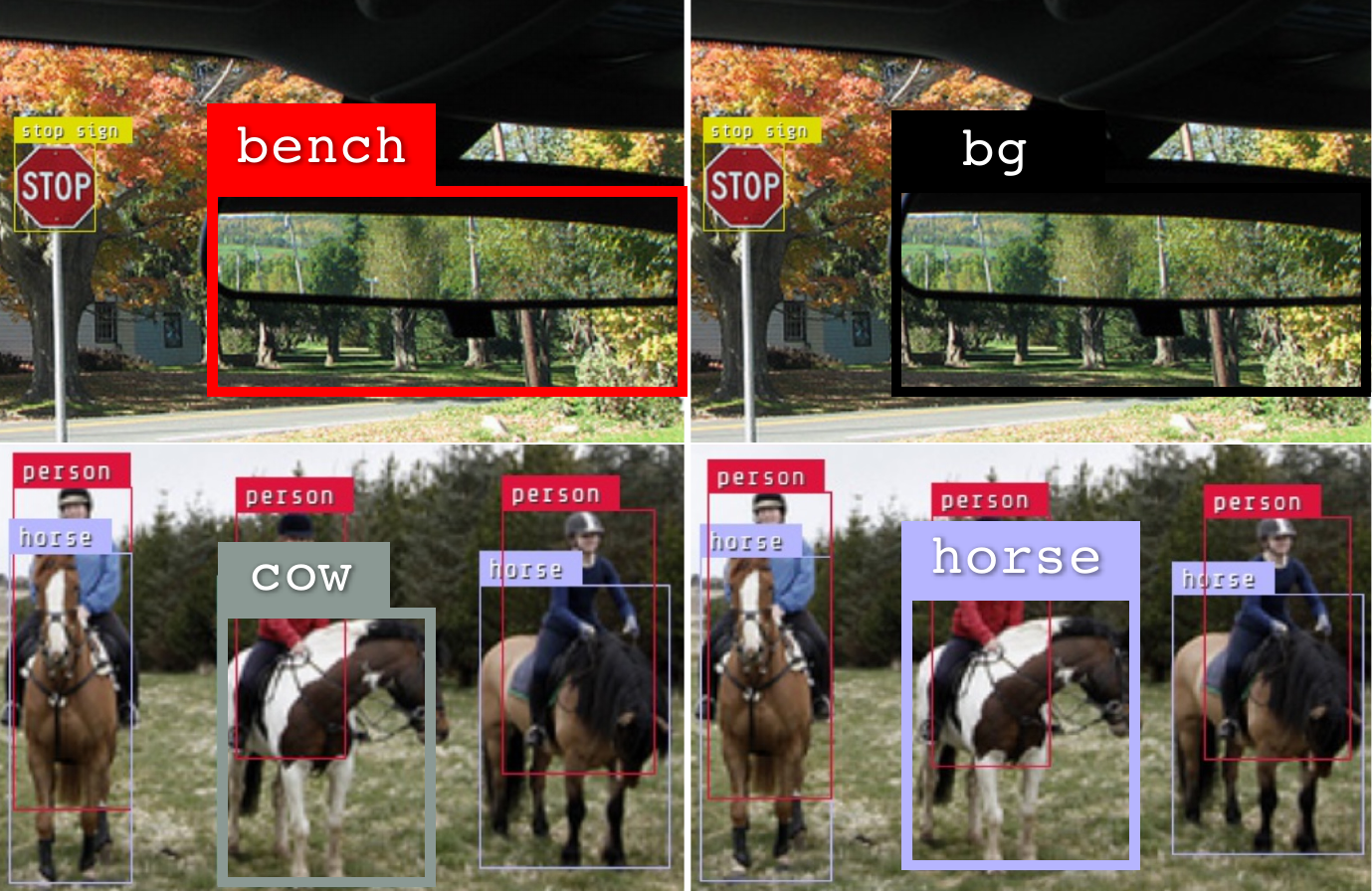}
   \hspace{1em}
   \includegraphics[align=c,width=0.6\textwidth]{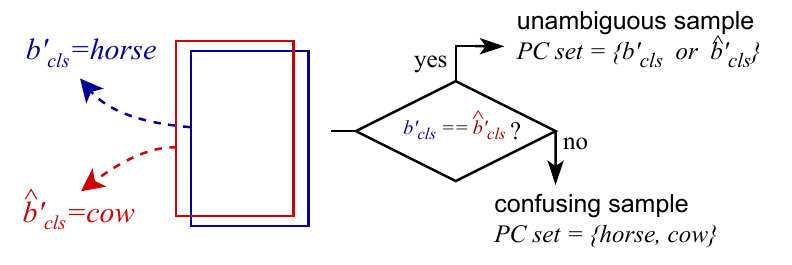}
   \vspace{-0.5em}
   \caption{Left: Examples of the confusing samples found by the temporal stability verification. Right: By comparing the current predictions (blue) with the last predictions (red) of the sample, the PC set should be \{horse, cow\}.}
   \label{fig:pc-example}
\vspace{-2em}
\end{figure}

\noindent It is observed that the pseudo label set $\mathcal{B}^{'}$ of an image varies at different training iteration steps~\cite{Qize_2021_Int}. As shown in~\cref{fig:pc-example}, by comparing $\mathcal{B}^{'}$ at different training steps, those changed pseudo labels reveal the potential categories. We select $\mathcal{B}^{'}$ of the current iteration step and $\mathcal{B}^{'}_{last}$ generated when the model viewed the current image the last time for the comparison.  Specifically, for one pseudo label $b^{'} \in \mathcal{B}^{'} \cup \mathcal{B}^{'}_{last}$, if there is any pseudo label $\hat b^{'} \in \mathcal{B}^{'} \cup \mathcal{B}^{'}_{last}$ that is close to $b^{'}$ and $b^{'}_{cls} = \hat b^{'}_{cls}$, no confusion will be recorded. Otherwise, if there is a nearby $\hat b^{'}$ but $b^{'}_{cls} \ne \hat b^{'}_{cls}$, $PC set =\{b^{'}_{cls}, \hat b^{'}_{cls}\}$. No nearby pseudo label box means that the confusing category is `background', thus $PC set =\{b^{'}_{cls}, bg\}$. The IoU is adopted to determine whether two pseudo bounding boxes are close enough or not.

\vspace{0.5em}\noindent\textit{b. Cross-model verification}

\noindent Inspired by the Co-training algorithm~\cite{Avrim_1998_Com}, comparing the predictions of two conditionally independent models for the same sample can also help to discover the potential category. Two detectors are initialised with different initial parameters. The orders of the training data for these two detectors are also different, ensuring that they will not collapse onto each other. For a pseudo bounding box $b^{'}_1$ from the detector-1, we use the detector-2 to re-determine the object in the region of $b^{'}_1$. The decision of the detector-2 is represented by $b^{'}_2$. Similar to the temporal stability verification, if $b^{'}_{cls_1}=b^{'}_{cls_2}$, the PC set contains only one category $b^{'}_{cls_1}$ (or $b^{'}_{cls_2}$). Otherwise, $PC set=\{b^{'}_{cls_1}, b^{'}_{cls_2}\}$.

In summary, for the classification training, If the size of the PC set is not equal to 1, it means that this sample should be a confusing sample. The VC loss will take over the training of this confusing sample. The standard CE loss is used for unambiguous samples.

\subsection{Localisation Loss}
Since the classification confidence score is not qualified to indicate the location quality of pseudo labels, some of the previous works~\cite{Yen-Cheng_2021_Unb,Kihyuk_2020_SATC} disabled the localisation loss of unlabelled data. We find that the creating method for the PC set can also measure the location quality of pseudo labels. When we create the PC set for a pseudo box $b$, we evaluate its location shift with the nearby box $\hat b$. We propose to decouple the horizontal and vertical boundary quality instead of using the IoU as a comprehensive metric to filter out the whole bounding box with low IoU value. The reason is that the IoU value can be affected by one biased boundary, even if the remaining boundaries are good. The horizontal quality flag $q_{hor}$ is calculated as:
\begin{equation}
    \label{eq:locflag}
    q_{hor}=\left\{
        \begin{array}{llll}
        1, & \frac{(x_1-\hat x_1)}{w} < t_{loc} & \& & \frac{(x_2-\hat x_2)}{w} < t_{loc}\\
        0, & otherwise & & \\
        \end{array} \right.,
\end{equation} where $x_1,x_2,\hat x_1,\hat x_2$ are the coordinates of the left and right boundary of the pseudo box $b$ and the nearby box $\hat b$, $w$ is the width of $b$, $t_{loc}$ is the threshold for high-quality boundaries. $q_{ver}$ is calculated in the same way.

The decoupling allows high-quality boundaries to contribute to the localisation training. For example, the regression of the left and right boundary can be trained when the horizontal boundary quality is satisfied, even the top and bottom boundaries are biased. The localisation loss of four Smooth-L1~\cite{Ross_2015_Fas} terms:
\begin{equation}
    \label{eq:locloss}
    \mathcal{L}_{reg^*}=q_{hor}\mathcal{L}^x + q_{ver}\mathcal{L}^y + q_{hor}\mathcal{L}^w + q_{ver}\mathcal{L}^h.
\end{equation}

In summary, the overall loss function is:

\begin{equation}
    \label{eq:overall loss}
    \mathcal{L}= \mathcal{L}_{CE}(\mathcal{D}_l) + \mathcal{L}_{reg}(\mathcal{D}_l) + \lambda(\mathcal{L}_{VC}(\mathcal{D}_u) + \mathcal{L}_{reg^*}(\mathcal{D}_u)),
\end{equation}where $\lambda$ is the weight for the loss of the unlabelled data. In the implementation, we also use the VC loss function to handle the unambiguous samples by only ignoring the index of the virtual category.

\section{Experiments}

\subsection{Datasets and Evaluation Protocol}
To evaluate the proposed method, we assess it on two well-known object detection benchmark datasets -- MS COCO~\cite{Tsung-Yi_2014_Mic} and Pascal VOC~\cite{Mark_2015_The}. Following the mainstream evaluation setting, we use the subset index provided by Unbiased Teacher~\cite{Yen-Cheng_2021_Unb} to split the \textit{train set} across five different labelled ratios: 0.5\%, 1\%, 2\%, 5\% and 10\% (each ratio are with five random seeds to get the averaged mAP). We also report the performance on Pascal VOC with \textit{VOC07-trainval} as the labelled subset and \textit{VOC12-trainval} as the unlabelled subset. The performance is evaluated on \textit{VOC07-test}. The evaluation metric of all the experiments reported in this paper is mAP calculated via the COCO evaluation kit~\cite{coco_eval_kit}.

\subsection{Implementation}
The proposed method is implemented in PyTorch framework~\cite{Adam_2019_PyT}. Following the mainstream choice of the community, we adopted Faster-RCNN~\cite{Shaoqing_2017_Fas} with FPN~\cite{Tsung-Yi_2017_Fea} and ResNet-50~\cite{Kaiming_2016_Dee} as the object detector. The weight term in Focal Loss~\cite{Tsung-Yi_2020_Foc} is also integrated into our VC loss to address the class imbalance issue.

The training is operated on 8 GPUs with batch size 1/4 per GPU for labelled/unlabelled data. The training iteration is 180k including a warmup of 2k steps with only the labelled subset. The optimiser is SGD with a constant $lr=0.01$. The default scaling factor for the virtual weight is a constant 3.5. The location quality threshold $t_{loc}$ in $\mathcal{L}_{reg^*}$ is 0.05. The default PC set creating method is the temporal stability verification. More details are introduced in the supplementary. The code can be found at: \href{https://github.com/GeoffreyChen777/VC}{https://github.com/GeoffreyChen777/VC}.

\subsection{Performance}
\noindent\textbf{MS COCO} We first evaluate our method on MS COCO with five label ratios. The 5 seeds averaged results are reported in~\cref{tab:coco}. `Ours' is our method with exactly the same settings of the solid baseline Unbiased Teacher for the sake of fairness. `Ours*' is obtained with some training tricks adopted by Soft Teacher. Results with $\dag$ are obtained from the available official code. The significant improvements can be summarised as follows:

\noindent 1) Compared with the supervised baseline, the mAP increases dramatically after training with the unlabelled data via our method.

\noindent 2) Our method outperforms other state-of-the-art semi-supervised detectors on all the label ratios by a significant margin. The mAP of our method at a small label ratio is close to or even exceeds the mAP of some methods at a large ratio.

\vspace{1em}\noindent\textbf{Pascal VOC} We also evaluate our method with VOC07 as the labelled subset and VOC12 and COCO* as the unlabelled subsets. We collect the images that contain objects in VOC predefined categories from MS COCO to build a subset \textit{COCO*}. The results are presented in~\cref{tab:voc}. Since the source codes of some methods are unavailable, the evaluation styles they used is unclear. \textit{Usually, the results based on the VOC-style AP are higher.} Thus, we evaluate our method with COCO-style mAP and VOC-style AP for the sake of fairness. Our method presents the best performance on these two unlabelled data subsets. Since VOC07 consists of more than 5K labelled images, and it is a relatively easy dataset,~\cref{tab:voc} indicates that our method can effectively further improve the performance, even if there is already sufficient labelled data. The similar conclusion can be drawn when we use the unlabelled COCO-additional and fully-labelled COCO-standard to validate our method (analysised in the supplementary).

\begin{table*}[t]
\centering
\renewcommand\arraystretch{0.6}
\scriptsize
\caption{The performance on MS COCO with different label ratios. Results with $\dag$ are obtained from the available official code. Ours is the results of the exactly same settings of Unbiased Teacher for the sake of fairness. Ours* is the results obtained by using some training tricks in Soft Teacher.}
\begin{tabular}{p{0.26\textwidth}|P{0.13\textwidth}|P{0.13\textwidth}|P{0.13\textwidth}|P{0.13\textwidth}|P{0.13\textwidth}}

\toprule
COCO label ratio &  0.5\% & 1\% & 2\% & 5\% & 10\% \\
\midrule
Supervised & 6.83 & 9.05 & 12.70 & 18.47 & 23.86\\
\midrule
CSD~\cite{Jisoo_2019_Con} & 7.41 & 10.51 & 13.93 & 18.63 & 22.46 \\
STAC~\cite{Kihyuk_2020_SATC} & 9.78 & 13.97 & 18.25 & 24.38 & 28.64 \\
Instant Teaching~\cite{Qiang_2021_Ins} & - & 18.05 & 22.45 & 26.75 & 30.40 \\
Interactive~\cite{Qize_2021_Int} & - & 18.88 & 22.43 & 26.37 & 30.53 \\
Humble Teacher~\cite{Yihe_2021_Hum} & - & 16.96 & 21.72 & 27.70 & 31.61 \\
Combating Noise~\cite{Zhenyu_2021_Com} & - & 18.41 & 24.00 & 28.96 & 32.43 \\
Soft Teacher~\cite{Mengde_2021_End} & 15.04\dag & 20.46 & 25.93\dag & 30.74 & 34.04 \\
Unbiased Teacher~\cite{Yen-Cheng_2021_Unb} & 16.94 & 20.75 & 24.30 & 28.27 & 31.50 \\
\midrule
Ours & 18.12 & 21.61 & 25.84 & 30.31 & 33.45 \\
Ours* & \textbf{19.46} & \textbf{23.86} & \textbf{27.70} & \textbf{32.05} & \textbf{34.82} \\
\bottomrule
\end{tabular}
\label{tab:coco}
\vspace{-1.2em}
\end{table*}

\begin{table}[t]
\vspace{-1em}
\centering
\scriptsize
\renewcommand\arraystretch{0.7}
\caption{Experiment results on VOC. $\mathcal{D}^l$ and $\mathcal{D}^u$ are the labelled and unlabelled subset choices. COCO* consists of the images from COCO that contains objects in VOC categories. Numbers in () are obtained with the VOC-style AP.}
\begin{tabular}{m{0.25\columnwidth}|m{0.05\columnwidth}|m{0.3\columnwidth}<{\centering}|m{0.3\columnwidth}<{\centering}}

\toprule
\multirow{2}*{Method} & $\mathcal{D}^l$ & \multicolumn{2}{c}{VOC07}\\
 & $\mathcal{D}^u$ & VOC12 & VOC12 + COCO* \\
\midrule
\multicolumn{2}{l|}{STAC~\cite{Kihyuk_2020_SATC}} & 44.64 & 46.01\\
\multicolumn{2}{l|}{Instant Teaching~\cite{Qiang_2021_Ins}}& 50.00 & 50.80 \\
\multicolumn{2}{l|}{Interactive~\cite{Qize_2021_Int}} & 46.23 & 49.59\\
\multicolumn{2}{l|}{Humble Teacher~\cite{Yihe_2021_Hum}} & 53.04 & 54.41 \\
\multicolumn{2}{l|}{Combating Noise~\cite{Zhenyu_2021_Com}} & 49.30 & 50.20  \\
\multicolumn{2}{l|}{Unbiased Teacher~\cite{Yen-Cheng_2021_Unb}} & 48.69 & 50.34  \\
\midrule
\multicolumn{2}{l|}{Ours} & \textbf{50.40 (55.74)} & \textbf{51.44 (56.70)} \\
\bottomrule
\end{tabular}
\label{tab:voc}
\vspace{-1.5em}
\end{table}

\begin{figure}[t]
\vspace{-1em}
\centering
\captionsetup[subfigure]{}
\subfloat[]{
    \centering
    \includegraphics[width=0.3\textwidth,height=0.22\textwidth]{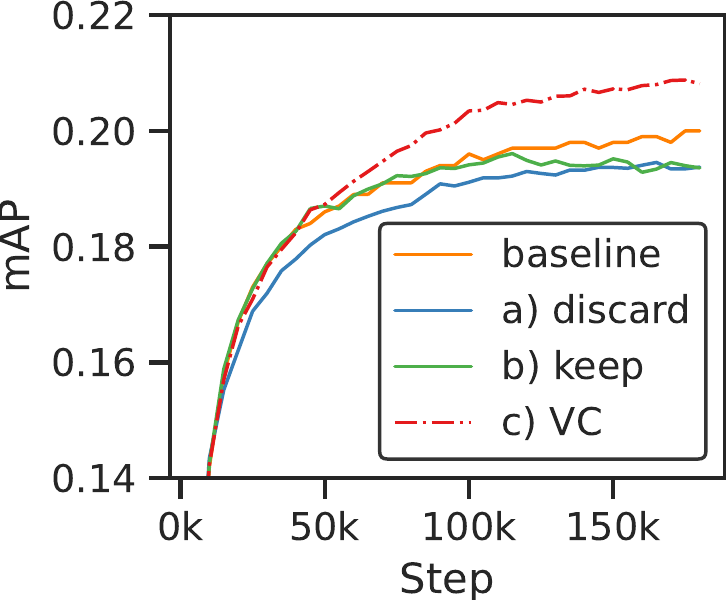}
    \label{fig:ablation-vc}
}
\subfloat[]{
    \centering
    \includegraphics[width=0.3\textwidth,height=0.22\textwidth]{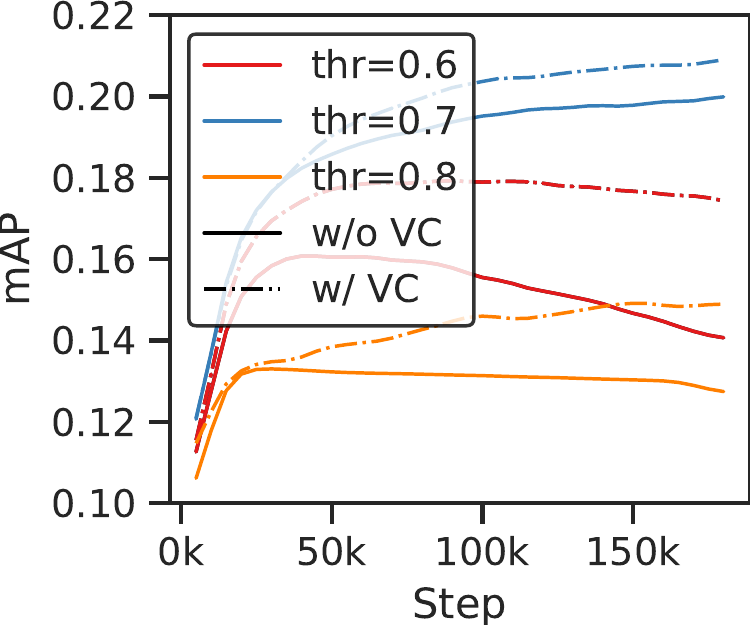}
    \label{fig:ablation-thr}
}
\subfloat[]{
    \centering
    \includegraphics[width=0.3\textwidth,height=0.22\textwidth]{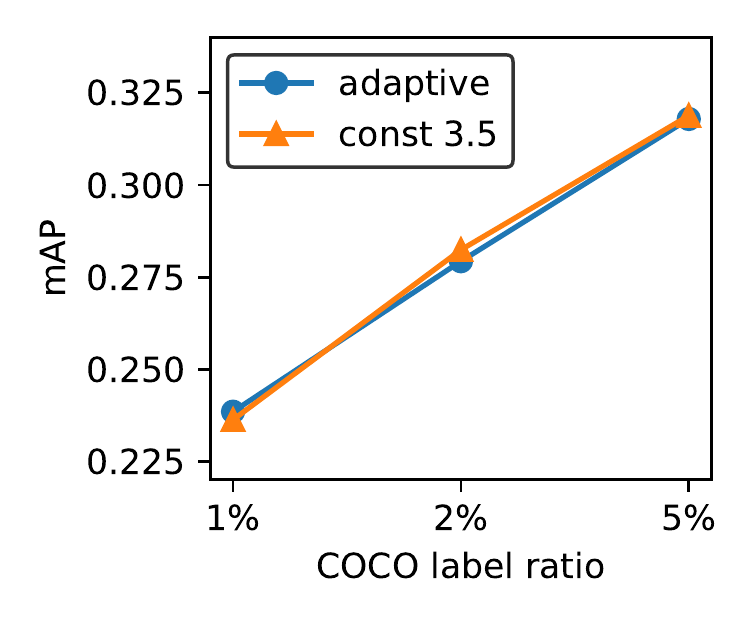}
    \label{fig:ablation-temp}
}
\\
\vspace{-1em}
\caption{a) Experiments of different strategies for dealing with confusing samples. As the pseudo labels of the valuable confusing samples are highly unreliable, it is not optimal to either discard or keep them. Our VC learning(dotdashed line) satisfies both demands, thereby resulting in a significant improvement. b) Experiments of different thresholds of the confidence score filtering w/ or w/o our VC learning. c) Comparison of the adaptive and the constant scaling factor.}
\vspace{-1.5em}
\end{figure}

\subsection{Analysis and Ablation Study}

In this section, we choose 1\% data of MS COCO as the labelled subset to analyse and validate our method in detail. All the experiments in this section are performed under the exactly same setting of the baseline model Unbiased Teacher except from the batch size. We adopt a smaller batch size to shorten the training time of each ablation study, therefore resulting in slightly decreased mAPs of all experiments compared to \cref{tab:coco}.

\subsubsection{Analysis of Virtual Category}
Here, we adopt the temporal stability verification to create the PC set for confusing samples. To analyse the effectiveness of the virtual category, we respectively report the mAP of the model under three policies: a) discarding all confusing samples (discard), b) retaining all potential labels for them (keep) and c) assigning our virtual category to replace the potential categories (VC). The baseline model is trained with vanilla pseudo labels (baseline) without the potential category discovery. As shown in~\cref{tab:ablation-vc}, both discarding and retaining policies decrease the mAP. By analysing the mAP during the whole training presented in~\cref{fig:ablation-vc}, we noticed that rejecting the confusing samples (blue line) results in a low mAP at the very beginning of the training. The reason is that using this policy discards some confusing samples with correct pseudo labels that the model desperately needs. Then, as shown by the green line in~\cref{fig:ablation-vc}, training with all PCs gives a small performance boost at the early stage of training because more underfitted samples are introduced to the model, but ends up with a low mAP. We believe this is due to the confirmation bias issue caused by incorrect pseudo labels that gradually hurts the performance. Our approach effectively resolves this conflict by providing a virtual category for the confusing sample.  The dotdashed line in~\cref{fig:ablation-vc} demonstrates that these confusing samples consistently benefit the model. The mAP sees a rise of 0.81 with our VC learning. The model with our VC learning exceeds the baseline early in the training and continues to lead until the end of training. 

In addition, as can be seen from~\cref{fig:ablation-thr}, we evaluate our VC learning with different thresholds (indicated by three colours) of the confidence score filtering adopted by our baseline model Unbiased Teacher. The confusing samples always exist, no matter whether the filtering mechanism is strict or not. The model with VC learning (dot-dashed lines) outperforms the baseline (solid lines) on three thresholds. Notably, the slump in the mAP disappears when $thr=0.6$, meaning that the confirmation bias has been effectively alleviated.

The constant and adaptive scaling factors of the virtual weight are compared in \cref{fig:ablation-temp}. The constant $3.5$ achieves a mAP which is comparable to the adaptive factor. Interestingly, we discover that $\frac{1}{ada. factor} \approx 3.5$.

\begin{table}[t]
\centering
\scriptsize
\begin{minipage}[t]{0.38\textwidth}
	\renewcommand\arraystretch{0.7}
	\centering
	\caption{Validation mAP with different strategies for dealing with the confusing samples.}
	\begin{tabular}{p{0.5\columnwidth}|P{0.45\columnwidth}}
		\toprule
		\hspace{1em}Strategies & mAP \\
		\midrule
		\hspace{1em}baseline & 20.00 \\
		\hspace{1em}a) discard & 19.37 \\
		\hspace{1em}b) keep & 19.36 \\
		\hspace{1em}c) VC & \textbf{20.81} \\
		\bottomrule
	\end{tabular}
	\label{tab:ablation-vc}
\end{minipage}
\hspace{1em}
\begin{minipage}[t]{0.56\textwidth}
	\renewcommand\arraystretch{0.6}
	\centering
	\caption{Ablation study on VC loss and modified localisation loss Reg* Loss. The creating method for the potential category set in VC learning is temporal stability verification.}
	\begin{tabular}{P{0.3\columnwidth}|P{0.3\columnwidth}|P{0.35\columnwidth}}
		\toprule
 		VC loss & Reg* Loss & mAP \\
		\midrule
 		&  & 20.00 \\
		$\surd$ & & 20.81 \\
		$\surd$ & $\surd$ & \textbf{20.94} \\
		\bottomrule
	\end{tabular}
	\label{tab:ablation-general}
\end{minipage}
\vspace{-2em}
\end{table}

\subsubsection{Ablation Study}
The overall ablation study is reported in~\cref{tab:ablation-general}. The model with VC learning and Reg* Loss performs favourably against the baseline model.

\vspace{0.5em}\noindent\textit{a. Creating methods of potential category set}

\noindent\cref{sec:pcset} explored two methods to create the PC set. We validate them and report the results in~\cref{tab:ablation-pc}. The cross-model verification achieves the best performance. The reason is that The cross-model verification is similar to the co-training technique which uses two independent models to provide pseudo labels for each other. it slightly alleviates the confirmation bias issue, thus resulting in additional improvement. As shown in~\cref{tab:ablation-pc}, the co-training can improve the mAP by 0.53 individually. In fairness to other methods without co-training, we use the temporal stability verification in all other experiments, although the cross-model verification performs better.

\begin{table}[t]
\centering
\scriptsize
\begin{minipage}[t]{0.38\textwidth}
	\renewcommand\arraystretch{0.9}
	\centering
	\caption{Ablation study of different image augmentation for virtual weights generation.}
	\begin{tabular}{p{0.6\columnwidth}|P{0.35\columnwidth}}
	\toprule
	Augmentations & mAP \\
	\midrule
	none & 20.80 \\
	flipping & \textbf{20.81} \\
	strong aug. & 20.70 \\
	\bottomrule
	\end{tabular}
	\label{tab:ablation-aug}
\end{minipage}
\hspace{1em}
\begin{minipage}[t]{0.56\textwidth}
	\renewcommand\arraystretch{0.6}
	\centering
	\caption{Ablation study of different methods for creating the PC. We also report the performance with only co-training techniques.}
	\begin{tabular}{p{0.6\columnwidth}|P{0.3\columnwidth}}
	\toprule
	Methods & mAP \\
	\midrule
	baseline & 20.00 \\
	Temporal & 20.81 \\
	Cross & \textbf{20.96} \\
	co-training w/o VC & 20.53 \\
	\bottomrule
	\end{tabular}
	\label{tab:ablation-pc}
\end{minipage}

\vspace{-2em}
\end{table}

\vspace{0.5em}\noindent\textit{b. Augmentation of the virtual weights}

\noindent We choose the feature vector from the teacher model to produce the virtual weight. It is natural to validate different augmentations for the input image of the teacher model to generate various virtual weights. We explore three different settings: no augmentation, horizontal flipping, and strong augmentation. The results are reported in~\cref{tab:ablation-aug}. No performance gap can be observed between no augmentation and only horizontal flip. Training with the virtual weight generated by the strong augmentation slightly degrades the mAP. The possible reason is that the strong augmentation, especially the cutout, significantly perturbs the input image. Thus, in the feature space, the direction of the virtual weight is far away from the weight vector of the GT category.

\vspace{0.5em}\noindent\textit{c. Hyperparameters}

\noindent The scaling factor ensures the norm of the virtual weight is in the similar range of the rest weight vectors. In addition to the adaptive version, we investigate the different constant scaling factor values here. The results (w/o $\mathcal{L}_{reg^*}$) are reported in~\cref{tab:ablation-hp-a}. $SC. = 3.5$ achieves the best mAP. The ablation study of the location quality threshold $t_{loc}$ in the $\mathcal{L}_{reg^*}$ are shown in~\cref{tab:ablation-hp-b}. A higher threshold will retain more unstable boundaries, thereby leading to worse performance.

\begin{table}[t]
\centering
\renewcommand\arraystretch{0.9}
\scriptsize
\caption{Ablation study of different hyperparameters. a) SC. is the scaling factor of the virtual weight. b) $t_{loc}$ is the threshold of the location quality.}
\begin{minipage}[t]{0.4\textwidth}
    \subfloat[]{
    \begin{tabular}{p{0.2\columnwidth}|P{0.2\columnwidth}|P{0.2\columnwidth}|P{0.2\columnwidth}}
        \toprule
        SC. & 2.5 & 3.5 & 4.5\\
        \midrule
        mAP & 20.63 & 20.81 & 20.65\\
        \bottomrule
    \end{tabular}
    \label{tab:ablation-hp-a}
    }
\end{minipage}
\begin{minipage}[t]{0.4\textwidth}
    \subfloat[]{
    \begin{tabular}{p{0.2\columnwidth}|P{0.2\columnwidth}|P{0.2\columnwidth}|P{0.2\columnwidth}}
    \toprule
    $t_{loc}$ & 0.03 & 0.05 & 0.1 \\
    \midrule
    mAP & 20.92 & 20.94 & 20.31\\
    \bottomrule
    \end{tabular}
    \label{tab:ablation-hp-b}
    }
    \label{tab:ablation-hp}
\end{minipage}
\vspace{-2em}
\end{table}

\section{Discussion}

In this section, we discuss the limitations and potential of VC learning. The VC loss takes over the optimisation of the confusing samples. Thus, higher quality of the PC set yields more improvement. The proposed creating methods are straightforward and easy to implement. Although a significant increment in mAP was observed already, there is still room to further boost the mAP with a better PC set. The mAP would increase from 20.81 to 24.10 if we were to use the ground truth to create the PC set under the setting of 1\% labelled COCO. Generating a better PC set can be worth exploring in the future. Moreover, by comparing the mAP gains of 10\% and other small label ratios, the improvement of our method is slightly lower, but still remains first. This phenomenon is expected and reasonable. On the one hand, more labelled data means a better baseline detector. Thus, the room between the baseline and the fully-supervised upper bound is smaller. On the other hand, less unlabelled data and a better detector indicate the confusing samples are fewer. At the extreme, with 100\% labelled data, our VC learning will be applied to no sample, thereby resulting in no improvement. Notably, this scenario is not the topic of this paper. We focus on the situation with very few labelled data. 

\section{Conclusion}
This paper proposed VC learning, which exploits the confusing underfitted unlabelled data. We provide a virtual category label to a sample if its pseudo label is unreliable. It allows the model to be safely trained with confusing data for further improvement to achieve state-of-the-art performance. To the best of our knowledge, VC learning is the first method to positively utilise confusing samples in SSOD rather than discarding them. It can also serve as a stepping stone to future work for the community of other semi-supervised learning tasks.

\bibliographystyle{splncs04}
\bibliography{egbib}
\end{document}